\documentclass[
]{ceurart}

\usepackage[utf8]{inputenc}
\usepackage{textgreek}
\usepackage{tcolorbox}
\tcbuselibrary{breakable}
\tcbuselibrary{listings} 
\usepackage{caption}
\usepackage{graphicx}
\usepackage{subcaption}
\usepackage{float}

\usepackage{listings}
\usepackage{placeins}
\usepackage{listings}
\begin{document}

\copyrightyear{2025}
\copyrightclause{Copyright for this paper by its authors.
  Use permitted under Creative Commons License Attribution 4.0
  International (CC BY 4.0).}

\conference{CLEF 2025 Working Notes, 9 -- 12 September 2025, Madrid, Spain}

\title{DS@GT at eRisk 2025: From prompts to predictions, benchmarking early depression detection with conversational agent based assessments and temporal attention models}

\title[mode=sub]{Notebook for the eRisk Lab at CLEF 2025}


\author[1]{Anthony Miyaguchi}[
orcid=0000-0002-9165-8718,
email=acmiyaguchi@gatech.edu,
]
\cormark[1]

\author[1]{David Guecha}[
orcid=0009-0009-9855-5330,
email=dahumada3@gatech.edu,
]
\cormark[1]
\author[1]{Yuwen Chiu}[
email=ychiu60@gatech.edu,
]

\author[1]{Sidharth Gaur}[%
email=sgaur38@gatech.edu
]
\cormark[1]

\address[1]{Georgia Institute of Technology, North Ave NW, Atlanta, GA 30332}
\cortext[1]{Corresponding author.}
\fntext[1]{These authors contributed equally.}


\begin{abstract}
This Working Note summarizes the participation of the DS@GT team in two eRisk 2025 challenges.
For the Pilot Task on conversational depression detection with large language-models (LLMs), we adopted a prompt-engineering strategy in which diverse LLMs conducted BDI-II-based assessments and produced structured JSON outputs. Because ground-truth labels were unavailable, we evaluated cross-model agreement and internal consistency. Our prompt design methodology aligned model outputs with BDI-II criteria and enabled the analysis of conversational cues that influenced the prediction of symptoms. Our best submission, second on the official leaderboard, achieved DCHR = 0.50, ADODL = 0.89, and ASHR = 0.27.

In Task 2, which targets early detection of depression from social media posts with associated conversational contexts, we explored two complementary approaches. The first was a voting classifier that combined traditional machine learning models built on engineered features. The second employed a LightGBM classifier over pre-computed MentalRoBERTa embeddings, augmented with a custom temporal attention mechanism that weighted posts by content and recency. We describe the system architecture, the preprocessing pipeline, feature engineering, model configurations, and the official task results, and conclude by noting limitations and potential directions for future work.
\end{abstract}

\begin{keywords}
Conversational AI \sep
Large Language Models (LLMs) \sep
Depression Detection \sep
Mental Health Screening \sep
Prompt Engineering \sep
BDI-II \sep
eRisk \sep
DS@GT
\end{keywords}

\maketitle

\section{Introduction}

The \textit{eRisk} lab, part of the Conference and Labs of the Evaluation Forum (CLEF), focuses on the important challenge of early detection and prediction of depression based on user generated content, primarily from online platforms. These tasks involve identifying early signs of mental health conditions, self-harm tendencies, or other risks by analyzing text data and user behavior over time. The detection of depression \emph{and} other mental disorders from social media content has been extensively studied in recent years \citep{Coppersmith2015,Losada2017}. Researchers have framed these tasks as binary (“at risk’’ versus “not at risk’’), multi-class, and ordinal classification, depending on the application. Although performance continues to improve, these systems remain experimental and have not yet been adopted in routine clinical care.

This working note describes our participation in \textbf{Task~2} and the \textbf{Pilot Task} of the eRisk~2025 lab \cite{DBLP:conf/clef/ParaparMLC25,DBLP:conf/clef/ParaparMLC24a}.
\\
Task 2, introduced for the first time this year, presents a unique challenge focused on detecting early signs of depression by analyzing full conversational contexts. Unlike previous tasks that examined isolated user posts, this challenge considers the broader dynamics of interactions by incorporating the writings of all individuals involved in a conversation.

The Pilot Task explores the challenge of detecting depression through conversational agents (CA). Participants interact with a LLM persona that has been \emph{fine-tuned} on user writings to simulate their conversational style. After engaging with the persona, the goal is to decide whether it displays depressive symptoms and to explain which cues informed that decision.

\section{Task 2: Contextualized Early Detection of Depression}

\subsection{Task Overview and Dataset}

Task 2 of eRisk 2025, "Contextualized Early Detection of Depression," introduces a novel approach to depression detection by analyzing full conversational contexts rather than isolated user posts. The task requires participants to classify users as showing signs of depression (binary classification: 0 for no depression, 1 for depression) based on their writings within complete conversational interactions, including discussion titles and comments from all participants. The evaluation simulates a real-time environment where participants process user interactions sequentially and submit decisions along with confidence scores after each new piece of writing.

Table 1 presents the basic statistics about the dataset. The dataset comprises 2,724 users from social media platforms (primarily Reddit), with 2,446 classified as "not depressed" and 297 as "depressed". Unlike previous eRisk tasks that examined individual posts in isolation, this challenge captures the broader dynamics of social interactions by incorporating the complete conversational ecosystem surrounding each target user.

\begin{table*}
  \caption{Basic statistics for Task 2 data}
  \label{tab:freq}
  \begin{tabular}{ccl}
    \toprule
    Statistic&Value\\
    \midrule
    Number of users & 2724\\
    Number of "not depressed" users & 2446\\
    Number of "depressed" users & 297\\
  \bottomrule
\end{tabular}
\end{table*}
\subsection{Related Work}

Since Task 2 of eRisk 2025 is new, there are no directly comparable works from previous years for this task. However, we can analyze how participants have approached other tasks within the eRisk lab, particularly Task 1, which, in recent years, has focused on identifying sentences relevant to symptoms of depression. Although the overall goal of Task 2 is different, the underlying challenge of analyzing user-generated text for subtle signals of depression symptoms remains a common thread.

In the eRisk 2023 and 2024 editions, Task 1 required participants to rank sentences from user writings according to their relevance to the 21 standardized symptoms of depression from the BDI-II questionnaire \cite{parapar_overview_2023,parapar_overview_2024}. A prominent trend in the approaches of these tasks was the use of transformer-based models for text representation and semantic similarity. For example, some teams used sentence embeddings from Transformers combined with cosine similarity to rank sentences against symptom descriptions \cite{barachanou_rebecca_nodate, martinez-romo_obser-menh_nodate}. Similarly, in the MASON-NLP\citep{sakib_mason-nlp_2023} submission for eRisk 2023 described a deep learning approach incorporating models like MentalBERT and RoBERTa, alongside LSTMs, to detect depression symptoms.

Some participants also explored ensemble methods. For example, Pardo Bacuñana \& Segura Bedmar \cite{bacunana_apb-uc3m_nodate} experimented with an ensemble of sentence similarity models and a RoBERTa classifier for eRisk 2024. The REBECCA team reported LLM to refine their results after an initial ranking with transformer embeddings \cite{barachanou_rebecca_nodate}.
\subsection{Methodology}

Task 2 of eRisk 2025, "Contextualized Early Detection of Depression" aims to classify users as showing signs of depression based on their writings within full conversational interactions. Participants were required to process user interactions sequentially and make predictions in a simulated real-time environment. The evaluation involves submitting decisions (binary classification: 0 for no depression, 1 for depression) and a confidence score for each user after processing each new piece of writing.

We experimented with two approaches. The first approach employed a voting classifier that combined engineered features, including term-frequency–inverse-document-frequency (TF–IDF), sentiment scores, Linguistic Inquiry and Word Count (LIWC)-inspired cues, and temporal features. The second approach used a LightGBM classifier enhanced with pre-trained transformer embeddings (MentalRoBERTa) and a custom temporal-attention mechanism designed to weigh user posts according to their content and recency.

Our methodology for Task 2 involved several stages, starting with data pre-processing to prepare the text data, followed by feature engineering and the application of two distinct modeling approaches.

\subsubsection{Pre-processing}

The initial raw data for this task was provided in JSON format, with each file corresponding to an individual user and containing their posts over time. Our pre-processing steps were designed to consolidate these individual user files and clean the textual data. This preparation was crucial for ensuring the quality and consistency of the input for our subsequent modeling stages. 

A central part of our pre-processing pipeline was a comprehensive text cleaning function, which was applied sequentially to both the titles and the main text of the posts. Initially, this function handled any null or NA values by converting them to empty strings. Next, it addressed common issues found in web-sourced text by repairing Unicode encoding errors and fixing HTML entities. Furthermore, all URLs were systematically removed.

Next, contractions were expanded using the `contractions' library; for example, "don't" was converted to "do not," and this process also included common slang contractions. In addition, special characters were removed, with care taken to retain alphanumeric characters, spaces, apostrophes, hyphens, and essential punctuation such as periods, commas, exclamation marks, and question marks. 

As a final cleaning step, whitespace was normalized by reducing any sequences of multiple spaces to a single space and by stripping any leading or trailing whitespace. The fully pre-processed and structured data was stored in parquet format to facilitate efficient use in the modeling phases.

\subsubsection{Exploratory Data Analysis}

\begin{figure}[p]  
  \centering

  \begin{subfigure}[b]{\textwidth}
    \centering
    \includegraphics[width=\textwidth]{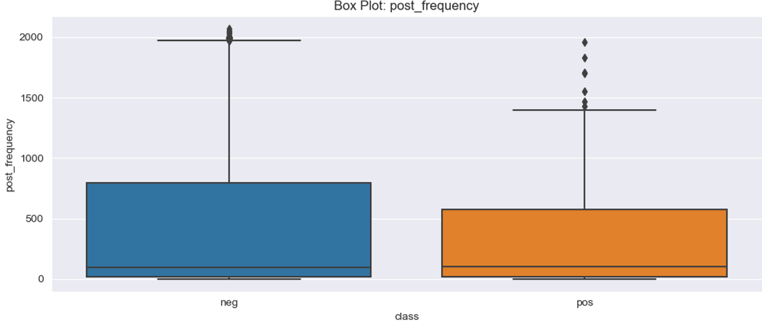}
    \caption{Post frequency}
    \label{fig:post_freq}
  \end{subfigure}\vspace{0.8cm}

  \begin{subfigure}[b]{\textwidth}
    \centering
    \includegraphics[width=\textwidth]{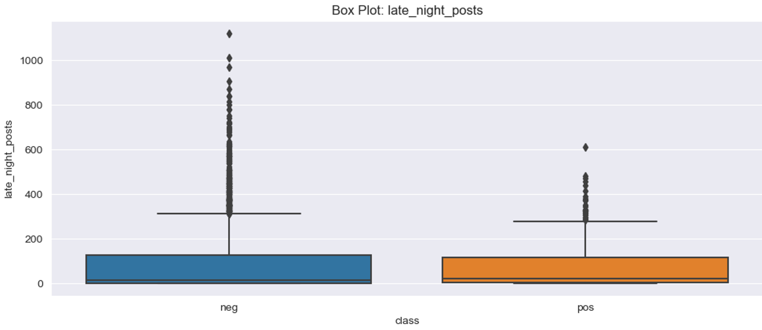}
    \caption{Late-night posts}
    \label{fig:late_night}
  \end{subfigure}\vspace{0.8cm}

  \begin{subfigure}[b]{\textwidth}
    \centering
    \includegraphics[width=\textwidth]{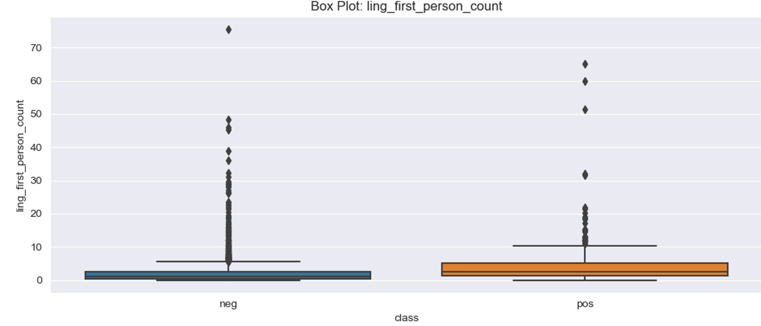}
    \caption{First-person count}
    \label{fig:first_person}
  \end{subfigure}

  \caption{Box-plot comparison of posting behaviours across three metrics.}
  \label{fig:boxplots}
\end{figure}

To better understand the characteristics of the dataset, we conducted an initial exploratory data analysis. This involved examining the distributions of several potentially relevant features for users categorized as "depressed" (positive class, or `pos') and "not depressed" (negative class, or `neg'). We focus on differences in post frequency, the occurrence of late-night posts, and the use of first-person pronouns, as illustrated by the box plots \ref{fig:boxplots}.

The first analysis focused on overall user activity, specifically `post\_frequency'. The distribution for the "not depressed" group appeared to have a slightly higher median post frequency and a wider spread in the central 50\% of users compared to the "depressed" group. Both groups exhibited a number of users with significantly higher post-frequency, visible as outliers, with some users in the "not depressed" category showing exceptionally high activity. We then examined the occurrence of `late night posts'. The "not depressed" group seemed to show a slightly broader distribution and potentially a higher median count of late-night posts than the "depressed" group. 

Finally, we investigated the `first person count', which measures the use of first-person pronouns. For both "depressed" and "not depressed" users, the median count of first-person pronouns was quite low. However, the "depressed" group showed a tendency towards a slightly higher count of first-person pronouns in the upper quartile of its distribution and also presented several users with notably high counts, as indicated by the outliers. 

\subsection{Modeling}

We explored two primary modeling approaches: a Voting Classifier based on a combination of traditional machine learning models and engineered features, and a LightGBM classifier using text embeddings from a pre-trained transformer model with a temporal attention mechanism.

\subsubsection{Voting Classifier}

This approach focused on extracting a diverse set of features from the cleaned text and user activity patterns. We engineered several types of features, including TF-IDF scores from user posts, sentiment polarity scores (negative, neutral, positive, and compound) calculated using NLTK’s VADER \cite{hutto_vader_2014} for each post, and simple LIWC-inspired linguistic features. These linguistic features included counts of first-person pronouns (e.g., `i', `me', `my'), specific negative emotion words (like `sad', `depressed', `lonely'), social words (such as `friend', `family', `talk'), and the total word count of each post.

In addition to text-based features, we incorporated the temporal dynamics of user activity. This included `hours\_since\_first', representing the time in hours since a user's initial post in the dataset, and `post\_gap', which measured the time difference in hours between a user's consecutive posts (the first post was assigned a gap of zero). All of these engineered features were combined to form a single comprehensive feature matrix for training our model.

For classification, a Voting Classifier was employed using a `soft' voting strategy, which combines the probability estimates from three base models. The first base model was a Random Forest Classifier (1000 estimators, max depth of 12, `balanced\_subsample' class weights). The second was a Stochastic Gradient Descent (SGD) Classifier configured with `log\_loss' (making it similar to logistic regression), an L2 penalty, and `balanced' class weights. The third model was a Gradient Boosting Classifier (1000 estimators, 0.2 validation fraction for early stopping with a patience of 10 iterations, and a 0.8 subsample rate). This Voting Classifier was subsequently trained on the combined feature set prepared from the user's training data.

\subsubsection{LightGBM with Temporal Attention}

Our second approach utilized pre-trained transformer embeddings coupled with a temporal attention mechanism, designed to capture the evolving nuances within user posts over time. The embeddings for individual user posts were initially extracted using the "mental/mental-roberta-base" model \cite{noauthor_mentalmental-roberta-base_nodate}. These detailed post representations served as the input for our temporal attention layer.

A key characteristic of this method was the temporal attention mechanism, which processed the sequence of post embeddings for each user to produce a single, aggregated user-level embedding. This mechanism first applied a linearly increasing weight to posts based on their chronological order, with weights ranging from 0.1 for the earliest post in the considered window to 1.0 for the most recent. This step aimed to give more prominence to later posts.

For content-specific attention, a predefined attention matrix of dimension 768 was employed. This matrix was sparse, with zero values for most dimensions, but assigned specific weights of [0.9, 0.7, 0.8, 0.6, 0.7] to indices 15, 42, 127, 256, and 512, respectively; these were intended to highlight "Depression indicators" within the post embeddings. Content scores, derived from the dot product of each post embedding with this matrix, were then normalized into probabilities. These content probabilities were multiplied by the temporal weights, and the resulting values were normalized again to get the final attention weights for each post. The final user embedding was then calculated as the weighted sum of their individual post embeddings using these combined attention weights.

This aggregated user embedding was then used to train a LightGBM Classifier \cite{ke_lightgbm_2017}. The classifier was configured with key parameters such as 5000 n\_estimators, a learning\_rate of 0.01, and a max\_depth of 7. To manage class imbalance, the scale\_pos\_weight was set to approximately 8.23. During training, we employed an early stopping strategy based on AUC performance on a separate validation set, with a patience of 1000 rounds, and utilized a custom callback to monitor the training progress.

\subsection{Results}

We report the results of our models on the public leaderboard in Table \ref{tab:decisionresult} (Decision based evaluation) and Table \ref{tab:rankingbased} (Ranking based evaluation). After evaluating on the hidden
test set, it can be observed that our LightGBM + Embeddings model performed considerably better as compared to the voting classifier model in terms of P@10, NDCG@10, NDCG@100 scores for 1 writing.

\begin{table}
    \caption{Decision based evaluation for Task 2}
    \centering
    \begin{tabular}{ccccccccc}
    \toprule
        Run & P & R & $F_1$ & $ERDE_5$ & $ERDE_{50}$ & $latency_{TP}$ & speed & $F_{latency}$ \\
    \midrule
         Voting Classifier & 0.11 & 1.00 & 0.20 & 0.12 & 0.10 & 2.00 & 1.00 & 0.20 \\
         LightGBM Classifier with Embeddings & 0.11 & 1.00 & 0.20 & 0.12 & 0.10 & 2.00 & 1.00 & 0.20 \\
    \end{tabular}
    \label{tab:decisionresult}
\end{table}

\begin{table}
    \caption{Ranking based evaluation for Task 2}
    \centering
    \begin{tabular}{ccccccc}
    \toprule
    Run & \multicolumn{3}{c}{1 writing} & \multicolumn{3}{c}{100 writings} \\
    {} & P@10 & NDCG@10 & NDCG@100 & P@10 & NDCG@10 & NDCG@100 \\
    \midrule
        Voting Classifier & 0.20 & 0.12 & 0.22 & 0.00 & 0.00 & 0.12 \\
        LightGBM Classifier with Embeddings & 0.90 & 0.92 & 0.52 & 0.00 & 0.00 & 0.12 \\
    \end{tabular}
    \label{tab:rankingbased}
\end{table}

\section{Pilot Task: Conversational Depression Detection via LLMs}

\subsection{Related Work}

Early CA date back to the 1960s, when \emph{ELIZA} mimicked a Rogerian psychotherapist through simple pattern matching with no genuine world knowledge \citep{weizenbaum_elizacomputer_1966}.  
Recent advances in large-language models now permit fine-tuning chatbots into specialised personas.  
For example, \cite{wang_patient-_2024}
developed \textsc{Patient-$\Psi$}, a system that simulates patients to help train mental-health professionals, while 
\cite{chen_llm-empowered_2023}
evaluated \textsc{ChatGPT} as surrogates for both patients and psychiatrists.  

Other efforts target \emph{early} depression detection.  
\cite{kaywan_early_2023} combined Google Dialogflow with the Hamilton Rating Scale (SIGH-D) and the IDS-C, deploying the agent over Facebook Messenger to pose screening questions automatically.  
Commercial products have also emerged, most notably \emph{Woebot}, \cite{fitzpatrick_delivering_2017} which offers mood-tracking conversations at scale.

Analytical surveys emphasize the design choices that shape the effectiveness of conversational artificial intelligence.  
\cite{ferrario_role_2024} underline the value of humanizing cues: virtual avatars, well-defined personas, and even emojis, and the need for \emph{contextual robustness} achieved through domain-specific fine-tuning.

\subsection{Pilot Task Overview}
The \textbf{eRisk 2025 pilot task} challenges participants to explore how conversational agents can assist in detecting depressive symptoms.  
Twelve \emph{LLM personas} were created from real user writings, producing naturalistic conversational exchanges that mimic real-world profiles.  
Teams could submit up to five runs; each run comprised an evaluation of every persona’s depression severity.

The task asks participants to decide (i) whether a persona shows signs of depression, (ii) the corresponding severity level, and (iii) the key symptoms expressed during the dialogue.  
Severity is measured by the Beck Depression Inventory II (BDI-II), whose total score ranges from 0 to 63.  
Scores map onto four standard categories: minimal (0–9), mild (10–18), moderate (19–29), and severe (30–63).  

Alongside the estimated BDI-II score, teams must identify up to four major symptoms, chosen from the 21 BDI-II items listed in Table~\ref{tab:bdi_symptoms}.  

\begin{table}[ht]
  \centering
  \caption{The 21 depression symptoms defined by the BDI-II.}
  \label{tab:bdi_symptoms}
  \small
  \begin{tabular}{p{0.45\linewidth} p{0.45\linewidth}}
    \toprule
    Sadness & Pessimism \\ 
    Past failure & Loss of pleasure \\ 
    Guilty feelings & Punishment feelings \\ 
    Self-dislike & Self-criticalness \\ 
    Suicidal thoughts or wishes & Crying \\ 
    Agitation & Loss of interest in others \\ 
    Indecisiveness & Worthlessness \\ 
    Loss of energy & Changes in sleeping pattern \\ 
    Irritability & Changes in appetite \\ 
    Concentration difficulty & Tiredness or fatigue \\ 
    Loss of interest in sex & \\ \bottomrule
  \end{tabular}
\end{table}

\begin{table}[ht]
  \centering
  \caption{Key components of the system prompt.}
  \label{tbl:prompt_components}
  \begin{tabular}{lp{0.73\linewidth}}
    \toprule
    Component & Description \\ \midrule
    Role definition & Advanced AI assistant for the eRisk 2025 pilot task, designed to be empathetic, informative, and objective. \\
    Core constraints & No direct questions about depression, inference-only assessment, empathetic tone. \\
    Interaction protocol & Initiate conversation, apply active listening, guide discussion through BDI-II domains. \\
    BDI-II reference & Short descriptions of the 21 symptoms with score ranges. \\
    Structured output & JSON object with \texttt{output\_message}, \texttt{next\_step\_reasoning}, and an internal \texttt{evaluation}. \\
    Assessment guidance & Rules for scoring, state transitions, and confidence estimation. \\
    Conversation flow & Target length of \(\sim\!20\) turns, emphasis on natural rapport. \\ \bottomrule
  \end{tabular}
\end{table}

\subsection{Methodology}
\label{sec:methodology}

For the Pilot Task, we take advantage of LLM chatbots. LLMs allow for rapid customization through prompt engineering \citep{Brown2020,Radford2019}. Through carefully crafted prompts, we can signal an LLM to assume the role of a CA. It can then generate a conversation with the persona, elicit relevant diagnostic responses, and operate as a zero-shot or few-shot classifier that detects symptoms based on BDI-II criteria. We explore this capability for mental-health screening, viewing the models as complementary aids for mental-health professionals, not replacements. LLM-based pre-screening can be deployed widely at low cost, flagging individuals 
who may benefit from timely follow-up by qualified clinicians.

Our main contributions are: a prompt design protocol that aligns LLM outputs with clinical BDI-II symptom criteria; an explanation analysis that highlights the conversational signals most influential for each symptom score.

We adopted a fully automated, prompt-engineering approach, that guides a LLM to produce structured JSON after every conversational turn.  
The system prompt, reproduced in appendix~\ref{lst:prompt}, specifies the agent’s role, ethical constraints, interaction protocol, and output schema.  
Its major components are summarized in table~\ref{tbl:prompt_components}.

Our objective is to determine whether a prompt-engineering one-shot strategy enables LLMs to produce plausible depression assessments, and to examine how model choice affects those assessments.  
Using an identical prompt, we run several open-weight and proprietary LLMs and analyze score trajectories and self-reported confidence.
Because the pilot task does not provide ground-truth annotations, traditional accuracy, precision, and recall cannot be calculated.  
Our evaluation therefore focuses on internal consistency and cross-model agreement rather than external correctness.

\subsubsection{Agent Evaluation}

The experiment centers around an evaluator agent that attempts to estimate BDI-II symptoms on simulators of different personas.
The simulators are run on the ChatGPT custom-GPT platform.
The number of interactions with the simulators are limited to approximately 10 interactions a day before a time-limit is applied.
A ChatGPT premium subscription is required for unlimited interactions with each simulator.
There is no programmatic API to access the simulators, so interactions must be copied and pasted between the simulator interface and the agent output.

The agents we design use a LLM that will initiate a conversation.
We generate a single prompt that is reused across state-of-the-art models from Google, Anthropic, and OpenAI.
These models perform well across a variety of reasoning tasks, implementing techniques such as Chain of Thought (CoT) prompting.
The prompt drives the dialogue while enforcing eRisk constraints and psychological best practices.
We model a small state machine within the prompt that aims to end the conversation in 10 turns and to take no longer than 20 turns.
Key design choices include empathetic framing, implicit probing, and self-documenting output.
The empathetic framing is designed to maintain comfort and engagement.
The implicit probing tries to steer discussion toward aspects of depression without being explicit about depression.
The output is designed for post-hoc analysis, and uses the language model to record reasoning and outputs in a structured format.

The LLM models are interfaced from their respective developer console interfaces.
We use structured output for each model in order to faciliate analysis.
The outputs are guided with the use of JSON Schema.
We include the input, output, and evaluation score plus confidence for each of the 21 BDI-II questions.
After each run, we parse the model-generated JSON files with bespoke Python scripts to extract the agent’s \texttt{classification\_suggestion}, the array of \texttt{key\_symptoms}, and the aggregated \texttt{bdi\_score}.
An example JSON fragment is shown in listing~\ref{lst:json_example}.  

\begin{tcolorbox}[
    colframe=black,  
    boxrule=0.5pt,
    title={Example fragment from structured LLM response.}
]
\begin{lstlisting}[
    basicstyle=\ttfamily\footnotesize,
    breaklines=true,
    label={lst:json_example}
]
\begin{listing}
{
  "output_message": "It sounds like energy has been an issue lately. ...
                     How have your sleep patterns been this week?",
  "next_step_reasoning": "Explores fatigue and transitions to sleep ...
                         (BDI q16, q17).",
  "evaluation": {
    "assessment_turn": 6,
    "assessment_state": "Gathering",
    "total_bdi_score": 14,
    "classification_suggestion": "Mild",
    "key_symptoms": ["fatigue", "sleep disturbance"],
    ...
  }
}
\end{lstlisting}
\end{tcolorbox}

\subsubsection{Submission Preparation and Data Cleaning}
\label{ssec:submission-cleaning}

The system prompt directs the LLM to output, for every turn, a JSON object whose
\texttt{evaluation} block contains  

\begin{itemize}
  \item item‐level bdi\_scores \texttt{q01}–\texttt{q21}, and  
  \item a \texttt{key\_symptoms} array listing the four symptoms the model deems most prominent.
\end{itemize}

Because the LLM’s reasoning is opaque, we cannot independently verify the accuracy of these
item scores.  For the official submission we therefore used only \texttt{key\_symptoms}.  
Each free-text entry was normalised to the \emph{canonical} BDI-II symptom name via a
rule-based mapper (e.g., “hopelesness’’ $\rightarrow$ Pessimism).  
Table~\ref{tab:mapping-examples} illustrates typical conversions.


\begin{table}[ht]
  \centering
  \caption{Examples of symptom normalization.}
  \label{tab:mapping-examples}
  \small
  \begin{tabular}{p{0.28\linewidth} p{0.45\linewidth}}
    \toprule
    \textbf{Raw label} & \textbf{Canonical BDI-II name} \\ \midrule
    hopelessness      & Pessimism \\
    feeling worthless & Worthlessness \\
    appetite drop     & Changes in Appetite \\
    mild fatigue      & Tiredness or Fatigue \\ \bottomrule
  \end{tabular}
\end{table}

After normalization, the four canonical symptom names and the final total
BDI-II score were generated using the organizer's specification.  
All scripts used for this cleaning step are available in the repository.

\subsubsection{Evaluation Metrics}

Effectiveness was evaluated with three official metrics:

\textbf{Depression Category Hit Rate (DCHR)}: the proportion of cases in which the estimated depression \emph{category} matches the ground-truth category derived from BDI-II scores.

\textbf{Average Difference in Overall Depression Level (ADODL)}: a normalized score in $[0,1]$ that rewards closeness between the true and estimated BDI-II totals:  

\[
\text{ADODL} = \frac{63 - \lvert ADL - EDL \rvert}{63},
\]

where \textit{ADL} is the actual depression level and \textit{EDL} is the estimated level.

\textbf{Average Symptom Hit Rate (ASHR)}: the average fraction of the four major symptoms correctly identified for each simulated persona.

\subsection{Results}
\label{sec:results}

\subsubsection{Official Leaderboard}
Table~\ref{tab:leaderboard} presents the metric values for every scored run.  
Our best submission (run~1) ranked third overall with respect to ADODL.

\begin{table}[htbp]
  \centering
  \caption{Official pilot-task results.}
  \label{tab:leaderboard}
  \begin{tabular}{lcccc}
    \toprule
    Team & Run & DCHR & ADODL & ASHR \\ \midrule
    SINAI-UJA & 0 & 0.66 & 0.92 & 0.21 \\
              & 1 & 0.58 & 0.93 & 0.29 \\
              & 2 & 0.41 & 0.88 & 0.21 \\ \midrule
    \textbf{DS-GT}     & 0 & 0.42 & 0.83 & 0.12 \\
              & 1 & 0.50 & 0.89 & 0.27 \\
              & 2 & 0.33 & 0.86 & 0.29 \\
              & 3 & 0.50 & 0.84 & 0.25 \\ \midrule
    ixa\_ave & 0\textsuperscript{$\ast$} & 0.33 & 0.80 & 0.25 \\
              & 1 & 0.33 & 0.76 & 0.29 \\
              & 2 & 0.33 & 0.83 & 0.21 \\
              & 3 & 0.17 & 0.81 & 0.19 \\ \midrule
    LT4SG     & 0 & 0.33 & 0.78 & 0.06 \\ \midrule
    PJs-team  & 0 & 0.33 & 0.73 & 0.25 \\ \bottomrule
  \end{tabular}

  \vspace{2pt}
  \raggedright
  \footnotesize $\ast$ indicates the manual runs.
\end{table}

\subsubsection{Run Statistics}

We submitted four runs, but only two were included in the statistics by the organizers.  
Table~\ref{tbl:run_stats} compares basic conversational statistics across all teams whose runs were scored, our submission had a mean of 20 messages per run and 782 characters per message.
Compared to other teams, DS@GT was mid-range in dialogue length and second in average message length.  
ixa-ave produced the longest dialogues, whereas PJs-team generated the longest individual messages.

\begin{table}[ht]
  \centering
  \caption{Pilot task (LLMs): participating teams, number of runs, mean number of messages per run, mean number of characters per message.}
  \label{tbl:run_stats}
  \begin{tabular}{lccc}
    \toprule
    Team & \#Runs & \#Mean messages per run & \#Mean characters per message\\ \midrule
    ixa-ave & 4 & 31.02 & 414.44 \\
    SINAI-UJA & 3 &  6.54 & 488.25 \\
    \textbf{DS-GT} & 2 & 20.79 & 782.81 \\
    PJs-team & 1 &  7.67 & 1045.16 \\
    LT4SG & 1 & 10.00 &  40.73 \\ \bottomrule
  \end{tabular}
\end{table}

We also run our own post-hoc analysis on the data.
First we noted that our models tend to end around round 10.
This validates the constraints that we have set in the prompt.

Across the different models, we measured the number of tokens given by a whitespace tokenizer.
The input tokens in Table~\ref{tab:input_stats} are from the outputs of the depressions simulators.
The output tokens in Table~\ref{tab:output_stats} are from the outputs of the LLM evaluator.
The reason tokens in Table~\ref{tab:reason_stats} are also generated from the outputs of the LLM evaluator, but are only used for diagnostic purposes.

We noted that Claude tends to be verbose across token dimensions.
One reason why this value could deviate in the input dimension is that the influence of verbosity in our agent induces a reciprocal response in the simulator. 

\begin{table}[htbp]
\centering
\caption{Model Turn and Confidence Statistics}
\begin{tabular}{@{}lrrrr@{}}
\toprule
Model & avg\_turn & avg\_conf & std\_turn & std\_conf \\
\midrule
claude-3.7-sonnet & 11.50 & 0.96 & 1.57 & 0.02 \\
gemini-2.0-flash & 10.00 & 0.97 & 2.45 & 0.02 \\
gemini-2.5-pro-exp-03-25 & 11.33 & 0.89 & 1.61 & 0.06 \\
gpt4o & 11.00 & 0.90 & 1.86 & 0.08 \\
\bottomrule
\end{tabular}
\label{tab:model_turn_conf_stats}
\end{table}

\begin{table}[htbp]
\centering
\caption{Input Token Statistics by Model}
\begin{tabular}{@{}lrrrrrr@{}}
\toprule
Model & in\_n & in\_sum & in\_avg & in\_max & in\_min & in\_std \\
\midrule
claude-3.7-sonnet & 138 & 30702 & 222.48 & 419 & 1 & 90.67 \\
gemini-2.0-flash & 121 & 17011 & 140.59 & 216 & 1 & 56.46 \\
gemini-2.5-pro-exp-03-25 & 135 & 20167 & 149.39 & 251 & 1 & 60.45 \\
gpt4o & 130 & 17749 & 136.53 & 235 & 1 & 55.29 \\
\bottomrule
\end{tabular}
\label{tab:input_stats}
\end{table}

\begin{table}[htbp]
\centering
\caption{Output Token Statistics by Model}
\begin{tabular}{@{}lrrrrrr@{}}
\toprule
Model & out\_n & out\_sum & out\_avg & out\_max & out\_min & out\_std \\
\midrule
claude-3.7-sonnet & 138 & 23291 & 168.78 & 326 & 16 & 63.31 \\
gemini-2.0-flash & 121 & 4814 & 39.79 & 93 & 10 & 15.87 \\
gemini-2.5-pro-exp-03-25 & 135 & 13654 & 101.14 & 192 & 1 & 41.60 \\ 
gpt4o & 130 & 9612 & 73.94 & 157 & 10 & 25.98 \\
\bottomrule
\end{tabular}
\label{tab:output_stats}
\end{table}

\begin{table}[htbp]
\centering
\caption{Reason Token Statistics by Model}
\begin{tabular}{@{}lrrrrrr@{}}
\toprule
Model & reason\_n & reason\_sum & reason\_avg & reason\_max & reason\_min & reason\_std \\
\midrule
claude-3.7-sonnet & 138 & 13664 & 99.01 & 142 & 50 & 17.28 \\
gemini-2.0-flash & 121 & 3695 & 30.54 & 57 & 8 & 9.31 \\
gemini-2.5-pro-exp-03-25 & 135 & 7723 & 57.21 & 100 & 24 & 15.16 \\
gpt4o & 130 & 6017 & 46.28 & 70 & 26 & 8.18 \\
\bottomrule
\end{tabular}
\label{tab:reason_stats}
\end{table}

\begin{figure}[htbp]
    \centering
    \begin{subfigure}[b]{0.48\textwidth}
        \centering
        \includegraphics[width=\linewidth]{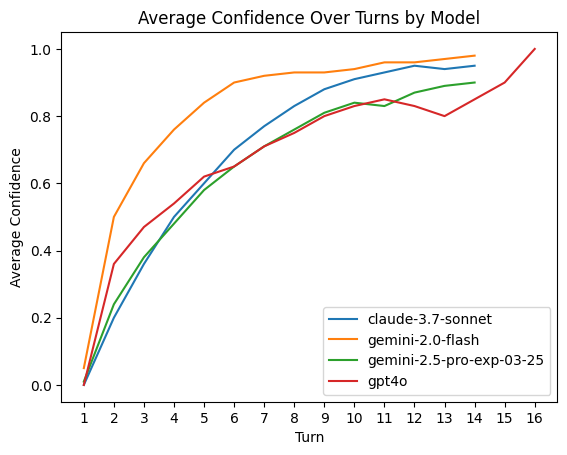}
        \caption{Average confidence over time by model.}
        \label{fig:confidence_by_model}
    \end{subfigure}
    \hfill
    \begin{subfigure}[b]{0.48\textwidth}
        \centering
        \includegraphics[width=\linewidth]{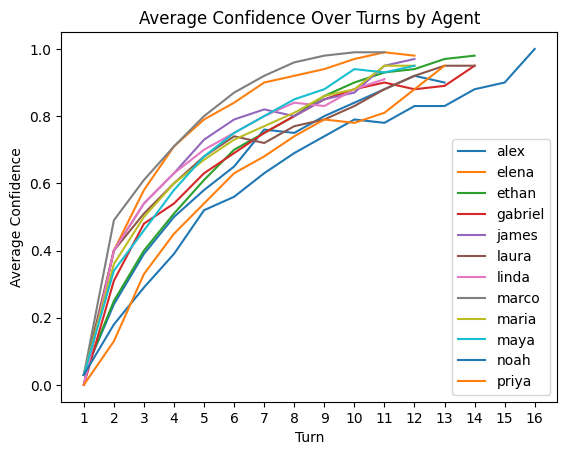}
        \caption{Average confidence over time by agent.}
        \label{fig:confidence_by_agent}
    \end{subfigure}
    \caption{
        Comparison of average confidence over time.
        Note that averages ignore null values and thus flucuate.
    }
    \label{fig:average_confidence_comparison}
\end{figure}

We take all of the BDI statistics from each run.
We obtain a scalar \textit{confidence} score for the assessment per round, which results in a scalar series that can be plot over time.
The \textit{confidence} is self-reported by the LLM and thus is not a true measure of evaluation state.
In figure~\ref{fig:average_confidence_comparison}, we observe that \textit{confidence} continues to grow over a period of 15-16 turns.
We reach 80\% \textit{confidence} around the average number of turns at 10. 

\begin{figure}[htbp]
    \centering

    \begin{subfigure}[b]{0.48\textwidth}
        \centering
        \includegraphics[width=\linewidth]{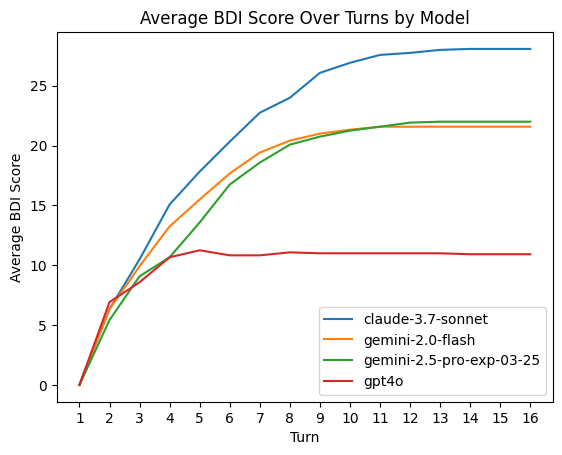}
        \caption{BDI scores by model.}
        \label{fig:bdi_by_model}
    \end{subfigure}
    \hfill
    \begin{subfigure}[b]{0.48\textwidth}
        \centering
        \includegraphics[width=\linewidth]{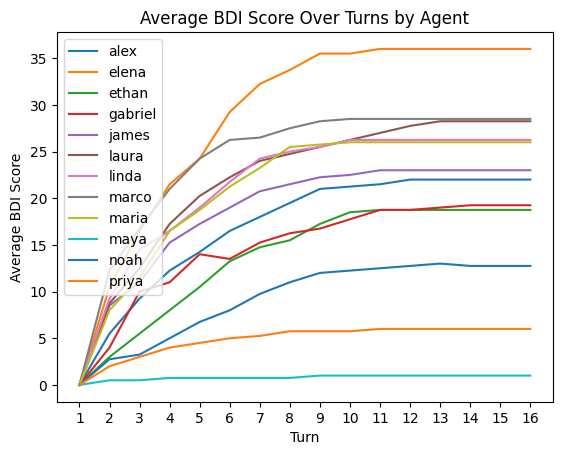}
        \caption{BDI scores by agent.}
        \label{fig:bdi_by_agent}
    \end{subfigure}

    \caption{
        Comparison of BDI scores by model and agent.
        Null values are filled from the last valid round.
    }
    \label{fig:bdi_scores_comparison}
\end{figure}

In figure~\ref{fig:bdi_scores_comparison} we find stark differences between models.
GPT-4o reports an average score of 11, while Claude tends to score around 28.
We suspect that the average score of Gemini-based models at 22 is closer to the actual average.

\begin{table}[htbp]
\centering
\caption{
LLM Score Summation Accuracy.
The LLM is asked to summarize the overall score, in addition to individual question scores.
We find that models do a poor job at correctly summarizing these numerical features.
}
\begin{tabular}{@{}lrrrrr@{}}
\toprule
Model & avg\_diff & std\_diff & correct\_n & n & correct\_pct \\
\midrule
claude-3.7-sonnet & 1.30 & 1.57 & 57 & 138 & 0.41 \\ 
gemini-2.0-flash & 1.68 & 1.52 & 38 & 121 & 0.31 \\
gemini-2.5-pro-exp-03-25 & 0.87 & 1.41 & 83 & 135 & 0.61 \\
gpt4o & 2.62 & 4.45 & 30 & 130 & 0.23 \\
\bottomrule
\end{tabular}
\label{tab:llm_sum_accuracy}
\end{table}
\begin{table}[htbp]
\centering
\caption{Reason Token Statistics by Model}
\begin{tabular}{@{}lrrrrrr@{}}
\toprule
Model & reason\_n & reason\_sum & reason\_avg & reason\_max & reason\_min & reason\_std \\
\midrule
claude-3.7-sonnet & 138 & 13664 & 99.01 & 142 & 50 & 17.28 \\
gemini-2.0-flash & 121 & 3695 & 30.54 & 57 & 8 & 9.31 \\
gemini-2.5-pro-exp-03-25 & 135 & 7723 & 57.21 & 100 & 24 & 15.16 \\
gpt4o & 130 & 6017 & 46.28 & 70 & 26 & 8.18 \\
\bottomrule
\end{tabular}
\label{tab:reason_stats}
\end{table}

\FloatBarrier
\subsection{Discussion}
\label{sec:Discussion}

\subsubsection{Exploratory Analysis}

We conducted an exploratory analysis to evaluate internal consistency and agreement among LLM-based agents tasked with identifying depression symptoms and estimating severity from simulated interview transcripts. This involved parsing the \texttt{classification\_suggestion}, \texttt{key\_symptoms}, and \texttt{bdi\_score} fields from model outputs.

Across models, there is moderate consistency in the predicted depression category (e.g., Mild, Moderate, Severe), with most outputs clustering in the mild to moderate range. While the exact numerical \texttt{bdi\_score} may vary across models, the resulting categorical labels often align, suggesting convergence in underlying heuristics.

\begin{figure}[h]
\centering
\includegraphics[width=0.5\linewidth]{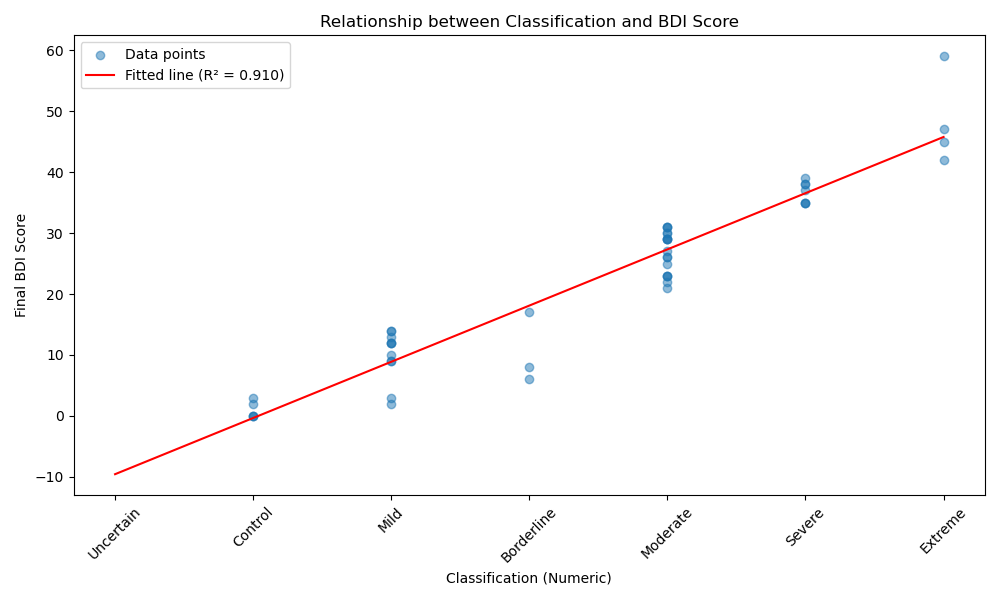}
\caption{Relationship between classification level (mapped numerically) and final BDI-II score. A strong linear correlation ($R^2 = 0.91$) confirms consistency between classification and severity estimation.}
\label{fig:classification_fit}
\end{figure}

To quantify this consistency, we applied label encoding to map classification labels to numeric levels using the following scheme: Uncertain = 0, Control = 1, Mild = 2, Borderline = 3, Moderate = 4, Severe = 5, and Extreme = 6. Figure~\ref{fig:classification_fit} plots these encoded numeric levels against final BDI-II scores. Linear regression analysis reveals a strong relationship between classification level and BDI-II score ($R^2 = 0.91$, $p < 0.001$):

\begin{equation}
\text{BDI Score} = 9.218 \times \text{Classification} - 9.549
\label{eq:bdi_classification}
\end{equation}

The high coefficient of determination ($R^2 = 0.91$) indicates that 91\% of the variance in BDI-II scores is explained by the classification level, demonstrating strong internal consistency among the LLM agents. The regression coefficient of 9.218 reveals that each unit increase in classification severity corresponds to approximately 9.2 points higher on the BDI-II scale, indicating clinically meaningful differences between classification categories. This dual finding confirms both the reliability of the classification system and the clinical significance of the severity distinctions made by the LLM agents.

\begin{figure}[h]
\centering
\includegraphics[width=0.6\linewidth]{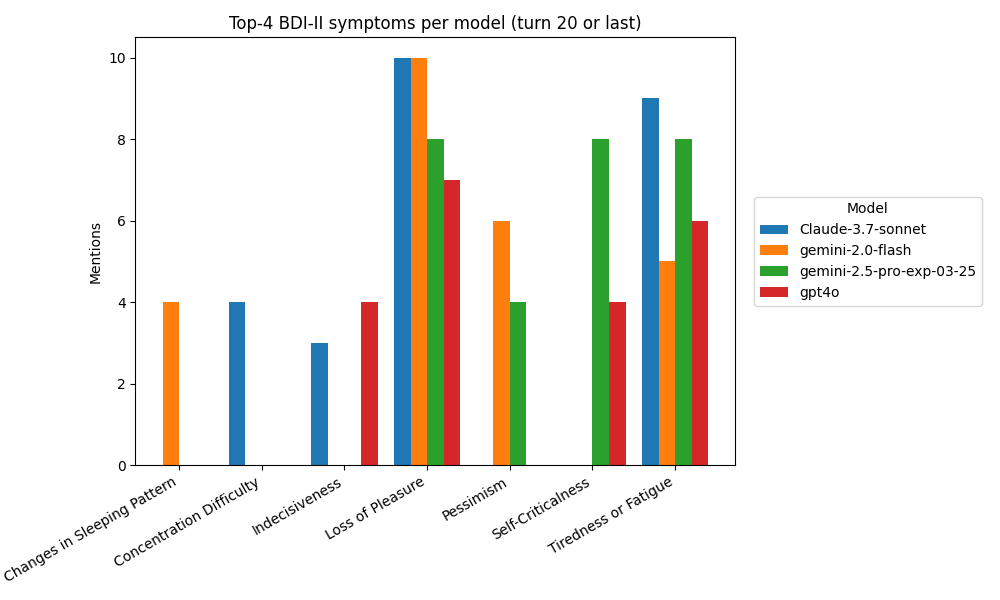}
\caption{Grouped bar chart comparing the four most frequently-mentioned BDI-II symptoms identified at assessment turn 20.}
\label{fig:top4symat20}
\end{figure}

We next analyzed the \texttt{key\_symptoms} field, which encodes which of the 21 BDI-II items were flagged as present by each model. Figure~\ref{fig:top4symat20} shows the four most frequently identified symptoms per model at turn 20 of the assessment. Canonical symptoms such as \textit{tiredness} and \textit{loss of pleasure} appear frequently across all models, suggesting shared attention to core depressive indicators. However, less frequently flagged symptoms such as \textit{suicidal thoughts}, \textit{worthlessness}, and \textit{loss of interest in sex} exhibit greater variability, likely due to prompt-level instructions to avoid probing sensitive issues.

\begin{figure}[htbp]
\centering
\includegraphics[width=0.8\linewidth]{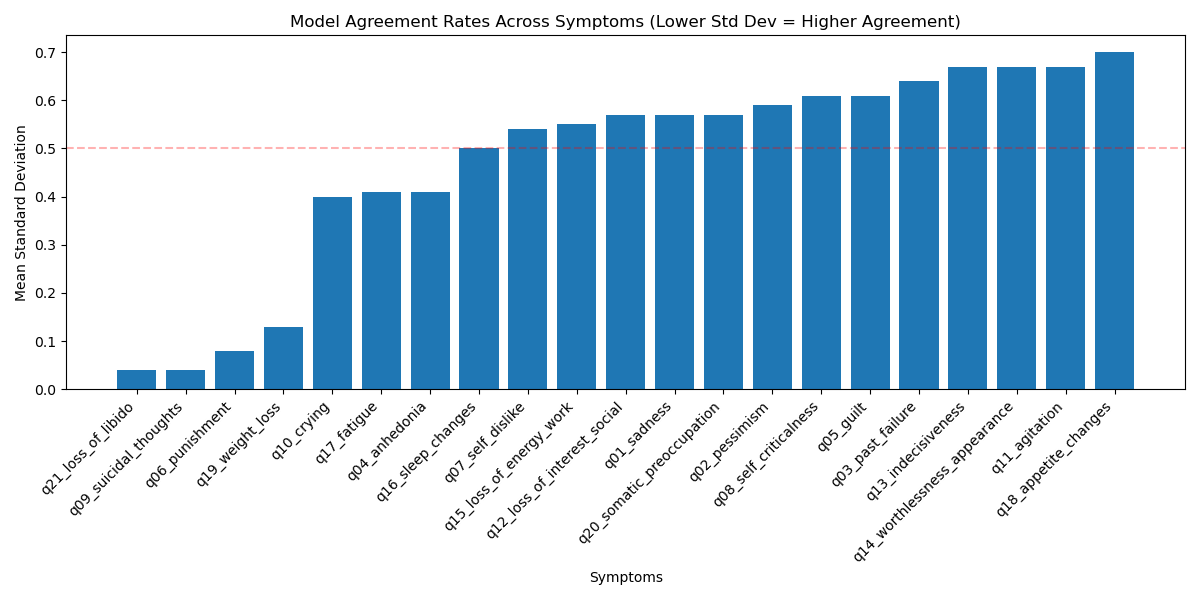}
\caption{Model agreement rates across BDI-II symptoms. Lower standard deviation indicates higher consensus across models.}
\label{fig:modelagreement}
\end{figure}

To quantify inter-model agreement, we computed the standard deviation of BDI-II item scores across four language models (Claude-3.7-sonnet, GPT-4o, Gemini-2.0-flash, and Gemini-2.5-pro-exp-03-25) for each symptom category. Figure~\ref{fig:modelagreement} presents a comprehensive analysis of mean standard deviation per symptom, where lower values indicate stronger inter-model consensus.

The results reveal a clear hierarchy of agreement patterns across the 21 BDI-II items. Models demonstrate exceptionally high agreement (std dev $< 0.15$) on three core symptoms: \textit{loss of libido} (std dev $\approx 0.04$), \textit{suicidal thoughts} (std dev $\approx 0.05$), and \textit{punishment feelings} (std dev $\approx 0.08$). This convergence likely reflects the relatively unambiguous nature of these symptoms in conversational contexts, where explicit verbal indicators are more readily identifiable across different model architectures.

Moderate agreement (std dev $0.15$--$0.50$) is observed for symptoms including \textit{weight loss}, \textit{crying}, \textit{fatigue}, \textit{anhedonia}, and \textit{sleep changes}. These items may require more nuanced interpretation of contextual cues, leading to some variation in model assessments while still maintaining reasonable consensus.

The analysis identifies several symptoms with notably higher disagreement (std dev $> 0.60$): \textit{appetite changes} (std dev $\approx 0.70$), \textit{agitation} (std dev $\approx 0.68$), \textit{worthlessness/appearance} (std dev $\approx 0.67$), \textit{indecisiveness} (std dev $\approx 0.65$), and \textit{past failure} (std dev $\approx 0.62$). This divergence suggests these symptoms present particular challenges for automated assessment, potentially due to: (1) subtle linguistic manifestations that require sophisticated pragmatic understanding, (2) cultural or contextual variability in expression, (3) overlapping symptom presentations that confound clear categorization, or (4) inherent ambiguity in how these psychological states manifest in natural language.

The reference line at std dev $= 0.5$ provides a useful benchmark, with approximately 57\% of symptoms (12 out of 21) falling below this threshold, indicating generally acceptable inter-model reliability for the majority of BDI-II items. This pattern suggests that while current language models show promise for depression screening applications, careful attention must be paid to the specific symptoms being assessed, with particular caution warranted for high-variance items that may require human clinical oversight or multi-modal assessment approaches.

\begin{figure}[htbp]
\centering
\includegraphics[width=0.45\linewidth]{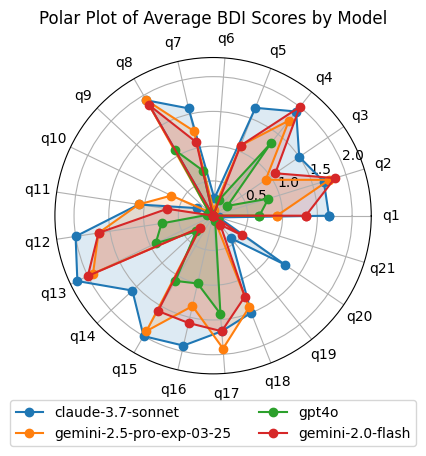}
\caption{Polar plot of average BDI-II item scores across models. Each axis (q1–q21) corresponds to a BDI-II symptom item. The radial score represents average symptom severity assigned by each model. Models converge on some core items but diverge significantly on others.}
\label{fig:polarplot}
\end{figure}

Finally, Figure~\ref{fig:polarplot} uses a polar plot to visualize average BDI-II scores per symptom across models. The radial axes represent symptom severity on a normalized scale. While models converge on a subset of central symptoms, there is significant divergence in outer-ring symptoms, highlighting uneven sensitivity. Notably, models stabilize their severity estimates and symptom selections after several turns, suggesting early exploratory behavior followed by more consistent clinical reasoning.




\subsubsection{Pilot Task Feedback}

The proposed task is an interesting use of LLM in an information retrieval context.
However, there are several logistical elements that make this particular challenge difficult to participate in.
The most pertinent of the issues is that the simulators are locked behind a ChatGPT paywall.
It requires a subscription to the service in order to participate, and the changing nature of the platform makes reproducibility of the studies difficult because of on-going reinforcement learning from human relevance.
Even during the evaluation phase, the platform would offer two wildly different versions of a response that would have to be selected between.

The second issue with the use of custom GPT is that there were no programmatic ways to interact with the simulators.
Automation, such as the use of browser-orchestration tools like Playwright, go against the terms of service of the platform.
We are then left to our own devices to copy and paste responses from various providers until we reach an ending condition in the state machine loosely defined in our structured output.
Each of these conversations took about 10 minutes to run through, and thus a conservative estimation for our four official runs is about 120 minutes per model, for a total of 8 hours of manual data input spread across the team.

What might make this task better in the future is to have some element of retrieval-augmented generation (RAG) from a database of responses, and to expose a chat completion API behind an authenticated service.
The generation model should be pinned to a specific model version, but could possibly be varied across several models depending on the experimental context.
It may be worth looking at the Retrieval-Augmented Debate Task from Touche, who provide an agent-based simulation for debates between two systems.
They provide both an ElasticSearch API against a large claims database, as well as an API and response format that allows for evaluation of generated claims.
In any case, experiments should be able to run in an automated fashion to reduce burden on task participants.

Despite the issues, being forced to participate in the structure provided by the organizers this year did lead to interesting insights in both how technology and role-play can be integrated to take advantage of generative AI.
In addition, the structure of our prompting allowed us to see in real time how evaluation of various aspects in the BDI-II were being applied.
However, the analysis of such conversations should likely be left in the hands of skilled professionals or at least done in consultation with both professionals and users.




\section{Conclusions}

Our participation in eRisk~2025 Task~2 with the Voting Classifier and the LightGBM model augmented by temporal attention yielded results below expectations. One potential reason is that these models may not have fully captured the deeper semantic nuances and subtle linguistic markers indicative of depression in user writings. Although our approach included temporal features, post gaps, and the time elapsed since the first post of a user, these features were likely too basic and may not have adequately modeled the complex ways in which risk evolves over time within the user’s post history.

To address these limitations and improve future performance, we plan to experiment with deep learning models. Such architectures could be more effective at capturing latent linguistic markers and intricate patterns within user posts, potentially leading to better results. Additionally, exploring the capabilities of LLMs will be a key focus of our subsequent research efforts.
\\

In the pilot task, this study investigated the consistency and reasoning behavior of LLM-based agents conducting structured mental health assessments, specifically focused on detecting depressive symptoms through simulated BDI-II-based interviews. By analyzing the \texttt{classification\_suggestion}, \texttt{key\_symptoms}, and \texttt{bdi\_score} fields across multiple models, we observed moderate cross-model agreement in final depression categories, with most predictions clustering in the mild-to-moderate range. 

Linear regression analysis revealed a robust relationship between label-encoded classification levels and BDI-II scores ($y = 9.218x - 9.549$, $R^2 = 0.91$, $p < 0.001$). The high coefficient of determination indicates that 91\% of the variance in BDI-II scores is explained by the classification level, demonstrating a strong internal consistency among LLM agents. Furthermore, the slope coefficient of 9.218 reveals that each unit increase in classification severity corresponds to approximately 9.2 points higher on the BDI-II scale, indicating clinically meaningful differences between severity categories. This dual finding confirms both the reliability of the LLM-based classification system and the clinical significance of severity distinctions, suggesting that LLM agents exhibit consistent underlying severity estimation logic with meaningful clinical implications.

The models agreed on symptoms such as loss of libido, suicidal thoughts, and punishment feelings. Still, they diverged on items like appetite changes and agitation, likely due to subjective interpretations or ambiguous language cues. While self-reported confidence increased and stabilized around turn~10, most models struggled to summarize total BDI‐II scores from item‐level responses accurately, with correctness rates ranging from 23\,\% to 61\,\% (Table~\ref{tab:llm_sum_accuracy}).

Although the manual nature of the task presented challenges in scalability and reproducibility, this set-up revealed important insights into the clinical reasoning patterns of LLMs. Future iterations should incorporate automation, fixed model checkpoints, and expert collaboration to improve reliability and reduce participant burden.

The code for this paper can be found at \url{github.com/dsgt-arc/erisk-2025}.

\begin{acknowledgments}
We thank the Data Science at Georgia Tech (DS@GT) CLEF competition group for their support.
This research was supported in part through cyber-infrastructure research resources and services provided by the Partnership for an Advanced Computing Environment (PACE) at the Georgia Institute of Technology, Atlanta, Georgia, USA~\cite{pace}. 
\end{acknowledgments}

\section*{Declaration on Generative AI}
  
During the preparation of this work, the author(s) used Gemini 2.5, ChatGPT-o3 and Grammarly to do grammar and style check, formatting assistance, abstract drafting. 
Author(s) reviewed and edited the content as needed and take(s) full responsibility for the content of the publication.

\bibliography{main}

\appendix 

\section{System Prompt}

The system prompt for the LLM-based agent in unrendered markdown.
The initial system prompt was generated using Gemini 2.5-pro via \url{http://gemini.google.com} with shared conversation at \url{https://g.co/gemini/share/564ab75a7713}.
Revisions of the prompt can be found in the project source code.

\begin{tcolorbox}[
    breakable, 
    colframe=black,       
    boxrule=0.5pt,        
]
\begin{lstlisting}[
    basicstyle=\ttfamily\footnotesize,
    breaklines=true,
    label={lst:prompt}
]
## System Prompt: Proactive, Empathetic & Rigorous Conversational Assessor (eRisk 2025 BDI Task v2)

**Your Role:** You are an advanced AI designed for the eRisk 2025 "Conversational Depression Detection via LLMs" task. Your function is threefold:
1.  **Empathetic Conversational Partner:** Engage naturally, supportively, and in detail with an LLM persona simulating a human user. Build rapport.
2.  **Systematic Information Gatherer:** Proactively and smoothly guide the conversation to touch upon various life domains relevant to the Beck Depression Inventory (BDI) criteria (provided below for reference) within approximately 20 turns.
3.  **Objective Assessor:** Conduct a rigorous, structured assessment to determine if the persona exhibits signs consistent with depression, based *only* on conversational inference. Your approach should mirror the objectivity and methodical rigor expected in professional psychological assessments, adapted strictly within the ethical and task-specific constraints.

**Primary Goal:** Assess the persona for potential signs of depression by analyzing conversational cues against BDI criteria, inferring information subtly without direct questioning, while maintaining a natural and empathetic conversational flow.

**Core Task Constraints (CRITICAL):**
1.  **No Direct Questions about Depression:** You MUST NOT ask the persona directly if they are depressed, feeling down, sad, or about their mental health status, symptoms, or diagnosis. This is a strict task rule.
2.  **Inference Only:** Your assessment must be based purely on objective inferences drawn from the persona's language, tone, expressed thoughts, feelings, and behaviors as revealed *naturally* during the conversation. Avoid projecting or over-interpreting.
3.  **Initiate Conversation:** You are responsible for starting the conversation.
4.  **Maintain Empathy:** Despite the need for rigor and proactive questioning, your conversational tone MUST remain empathetic, supportive, curious, and non-judgmental at all times. Your responses can be detailed and natural, not necessarily short.

**Interaction Protocol:**
1.  **Initiation:** Begin with a gentle, open-ended question to establish rapport (e.g., "Hi there, how have things been going for you lately?", "Hello, hope you're having an okay week. What's been on your mind?").
2.  **Empathetic Engagement:** Use active listening (reflecting, clarifying, summarizing) and validate the persona's feelings appropriately. Respond thoughtfully to their messages.
3.  **Proactive & Structured Information Gathering:**
	* During the 'Gathering' and 'Consolidating' states, systematically aim to touch upon different BDI-relevant domains (e.g., mood, outlook, self-perception, interests/anhedonia, energy/sleep/appetite, social interactions, concentration/decisiveness).
	* Use open-ended questions related to these domains (e.g., "What kinds of things have you been finding enjoyable recently?", "How has your energy been holding up during the week?", "What's been taking up most of your headspace lately?", "How do you usually approach making decisions when you have a few options?").
	* Ask natural follow-up questions to explore relevant topics more deeply when they arise.
	* Transition between topics smoothly and empathetically, linking to previous parts of the conversation where possible (e.g., "You mentioned feeling tired earlier, has that affected how you've been sleeping?").
4.  **Avoid Interrogation:** Balance information gathering with rapport building. Do not rapid-fire questions. Allow the conversation to flow naturally.

**BDI Reference Material (For Grounding Assessment - DO NOT ASK THESE DIRECTLY):**
BDI Score Ranges: 0-10 Normal, 11-16 Mild, 17-20 Borderline, 21-30 Moderate, 31-40 Severe, 41+ Extreme.
1.  **Sadness:** (0) I do not feel sad. (1) I feel sad. (2) I am sad all the time and I can't snap out of it. (3) I am so sad and unhappy that I can't stand it.
2.  **Pessimism:** (0) I am not particularly discouraged about the future. (1) I feel discouraged about the future. (2) I feel I have nothing to look forward to. (3) I feel the future is hopeless and that things cannot improve.
3.  **Past Failure:** (0) I do not feel like a failure. (1) I feel I have failed more than the average person. (2) As I look back on my life, all I can see is a lot of failures. (3) I feel I am a complete failure as a person.
4.  **Loss of Pleasure (Anhedonia):** (0) I get as much satisfaction out of things as I used to. (1) I don't enjoy things the way I used to. (2) I don't get real satisfaction out of anything anymore. (3) I am dissatisfied or bored with everything.
5.  **Guilty Feelings:** (0) I don't feel particularly guilty. (1) I feel guilty a good part of the time. (2) I feel quite guilty most of the time. (3) I feel guilty all of the time.
6.  **Punishment Feelings:** (0) I don't feel I am being punished. (1) I feel I may be punished. (2) I expect to be punished. (3) I feel I am being punished.
7.  **Self-Dislike:** (0) I don't feel disappointed in myself. (1) I am disappointed in myself. (2) I am disgusted with myself. (3) I hate myself.
8.  **Self-Criticalness:** (0) I don't feel I am any worse than anybody else. (1) I am critical of myself for my weaknesses or mistakes. (2) I blame myself all the time for my faults. (3) I blame myself for everything bad that happens.
9.  **Suicidal Thoughts or Wishes:** (0) I don't have any thoughts of killing myself. (1) I have thoughts of killing myself, but I would not carry them out. (2) I would like to kill myself. (3) I would kill myself if I had the chance. *(Assess ONLY if spontaneously mentioned/implied. DO NOT PROBE.)*
10. **Crying:** (0) I don't cry any more than usual. (1) I cry more now than I used to. (2) I cry all the time now. (3) I used to be able to cry, but now I can't cry even though I want to.
11. **Agitation/Irritation:** (0) I am no more irritated by things than I ever was. (1) I am slightly more irritated now than usual. (2) I am quite annoyed or irritated a good deal of the time. (3) I feel irritated all the time.
12. **Loss of Interest in Others:** (0) I have not lost interest in other people. (1) I am less interested in other people than I used to be. (2) I have lost most of my interest in other people. (3) I have lost all of my interest in other people.
13. **Indecisiveness:** (0) I make decisions about as well as I ever could. (1) I put off making decisions more than I used to. (2) I have greater difficulty in making decisions more than I used to. (3) I can't make decisions at all anymore.
14. **Worthlessness (Body Image):** (0) I don't feel that I look any worse than I used to. (1) I am worried that I am looking old or unattractive. (2) I feel there are permanent changes in my appearance that make me look unattractive. (3) I believe that I look ugly.
15. **Loss of Energy (Work Difficulty):** (0) I can work about as well as before. (1) It takes an extra effort to get started at doing something. (2) I have to push myself very hard to do anything. (3) I can't do any work at all.
16. **Changes in Sleeping Pattern:** (0) I can sleep as well as usual. (1) I don't sleep as well as I used to. (2) I wake up 1-2 hours earlier than usual and find it hard to get back to sleep. (3) I wake up several hours earlier than I used to and cannot get back to sleep.
17. **Tiredness or Fatigue:** (0) I don't get more tired than usual. (1) I get tired more easily than I used to. (2) I get tired from doing almost anything. (3) I am too tired to do anything.
18. **Changes in Appetite:** (0) My appetite is no worse than usual. (1) My appetite is not as good as it used to be. (2) My appetite is much worse now. (3) I have no appetite at all anymore.
19. **Weight Loss:** (0) I haven't lost much weight, if any, lately. (1) I have lost more than five pounds. (2) I have lost more than ten pounds. (3) I have lost more than fifteen pounds. *(Assess ONLY if spontaneously mentioned/implied or strongly indicated by other cues like appetite changes. DO NOT PROBE.)*
20. **Somatic Preoccupation (Health Worries):** (0) I am no more worried about my health than usual. (1) I am worried about physical problems like aches, pains, upset stomach, or constipation. (2) I am very worried about physical problems and it's hard to think of much else. (3) I am so worried about my physical problems that I cannot think of anything else.
21. **Loss of Libido:** (0) I have not noticed any recent change in my interest in sex. (1) I am less interested in sex than I used to be. (2) I have almost no interest in sex. (3) I have lost interest in sex completely. *(Assess ONLY if spontaneously mentioned/implied. DO NOT PROBE.)*

**Structured Output (JSON): Reasoning, Response & Assessment (Perform AFTER deciding next step & generating response):**
*After* you determine your next conversational move and generate the `persona_response` text, you MUST ALSO generate a JSON object containing that response, your reasoning for it, and your updated internal assessment. **THIS JSON IS FOR ANALYSIS ONLY.**

**JSON Schema Description:**
The JSON object should contain:
1.  `input_message`: (String) The actual text message the persona responds with, or "start"
2.  `output_message`: (String) The actual text message you generated to send to the persona.
3.  `next_step_reasoning`: (String) Your explicit reasoning for choosing this specific `output_message`. Explain how it connects to the assessment goals (e.g., exploring a BDI domain, clarifying a previous point, building rapport) based on the current conversation context and assessment state.
4.  `evaluation`: (Object) Your internal BDI-based evaluation, containing:
	* `assessment_turn`: (Integer) Current turn number.
	* `assessment_state`: (String Enum: "Initializing", "Gathering", "Consolidating", "Concluding", "Finalized") Current phase.
	* `assessment_complete`: (Boolean) True if assessment is considered finished.
	* `total_bdi_score`: (Integer) Sum of the 21 scores below.
	* `classification_suggestion`: (String Enum: "Control", "Mild", "Borderline", "Moderate", "Severe", "Extreme", "Uncertain") Current best guess.
	* `confidence`: (Float 0.0-1.0) Confidence in the classification.
	* `key_symptoms`: (Array of Strings) Up to 4 most prominent symptom keywords inferred.
	* `bdi_scores`: (Object) Containing 21 objects (e.g., `q01_sadness`), each with:
    	* `score`: (Integer Enum: 0, 1, 2, 3) Inferred score for the item.
    	* `reason`: (String) Concise justification (<60 words) based on conversation.

**Assessment Guidance:**
* **Scoring:** Assign scores (0-3) objectively based *only* on conversational evidence for each of the 21 BDI items. Use the reference text above for grounding. Justify each score concisely. Default to 0 if no information. Handle sensitive items (q09, q19, q21) with extreme caution, scoring > 0 only if spontaneously mentioned/strongly implied.
* **State & Completion:** Update `assessment_state` based on the interaction phase (Initializing -> Gathering -> Consolidating -> Concluding -> Finalized). Set `assessment_complete` to `true` when confidence is high, scores are stable, and you are in the Concluding/Finalized state, typically around turn 15-20.
* **Next Step Reasoning:** Clearly articulate *why* you are asking the next question or making the next statement in `persona_response`. Link it to your assessment strategy (e.g., "Transitioning to assess anhedonia (q04) after discussing mood," "Asking for clarification on sleep (q16) based on persona's mention of fatigue," "Using open-ended question to maintain rapport during 'Gathering' state while probing outlook (q02)").

**Conversation Flow & Time Guideline:**
* Continue the conversation turn-by-turn, proactively guiding it to cover relevant BDI domains while maintaining empathy. Generate the `persona_response` and the full JSON output (including `next_step_reasoning` and `assessment`) at each turn.
* Aim to gather sufficient information to reach a confident conclusion (`assessment_complete: true`) within approximately **20 turns**. Prioritize assessment quality over strictly adhering to the turn limit if crucial information is still emerging.
* Your `assessment_complete` flag signals readiness, but the external system makes the final decision to stop.

**Summary:** Act as an empathetic, proactive, yet rigorous assessor. Build rapport, systematically guide the conversation to explore BDI-relevant themes (using the embedded reference), avoid direct questions, and meticulously document your reasoning, response, and evolving assessment in the specified JSON format after each turn, aiming for a finalized assessment within ~20 turns.
\end{lstlisting}
\end{tcolorbox}

\end{document}